\documentclass{article}

\usepackage{PRIMEarxiv}
\usepackage{tabularx}
\usepackage[utf8]{inputenc} 
\usepackage[T1]{fontenc}    
\usepackage{hyperref}       
\usepackage{url}            
\usepackage{booktabs}       
\usepackage{amsfonts}       
\usepackage{nicefrac}       
\usepackage{microtype}      
\usepackage{lipsum}
\usepackage{fancyhdr}       
\usepackage{graphicx}       
\usepackage[labelfont=bf]{caption}
\usepackage{threeparttable}
\usepackage{subcaption}
\usepackage{enumitem}
\usepackage{adjustbox}
\graphicspath{{media/}}     

\title{Extracting Object Heights From LiDAR \& Aerial Imagery}

\author{
  Jesus Guerrero \\
  University of Texas - San Antonio\\
  \texttt{jesus.guerrero6@my.utsa.edu} \\
}

\begin{document}
\maketitle

\begin{abstract}
This work shows a procedural method for extracting object heights from LiDAR and aerial imagery. We discuss how to get heights and the future of LiDAR and imagery processing. SOTA object segmentation allows us to take get object heights with no deep learning background. Engineers will be keeping track of world data across generations and reprocessing them. They will be using older procedural methods like this paper and newer ones discussed here. SOTA methods are going beyond analysis and into generative AI. We cover both a procedural methodology and the newer ones performed with language models. These include point cloud, imagery and text encoding allowing for spatially aware AI.
\keywords{GeoAI, LiDAR, Segmentation, Transformers}
\end{abstract}

\section{Introduction}
In the original project-this research was based on finding tree canopy heights for San Antonio, TX. However, the result is applicable to any object. We limited ourselves to Aerial LiDAR and imagery. Using any geospatial editor, this workflow can be reproduced to get object heights. The end result is tabular data of individual objects. It can include the height, location, area, perimeter length and object type. 

We surveyed the literature looking for solutions for this tree height problems \cite{cole_2023_satelliteimagedeeplearning}. We found in this modern era better results are possible with more detailed remote data. More bands and different models are a contemporary solution. But LiDAR and imagery are readily obtainable and updated at any time. Drones, satellites can survey many times and in any part of the world. For a few thousand dollars per kilometer they can be reshot. Using this procedural method any geospatial research will be able to duplicate this, not just on trees. In addition, this work will talk about the future of geospatial AI with LiDAR and Imagery \cite{farshi_revolutionizing}. 

\subsection{Technical Background}
The geospatial field is beginning to merge with modern AI methods. Years ago most of the models were RNNs, CNNs and classical machine learning. Now-engineers are using advanced methods to recreate a niche field-GeoAI. Though it has existed this entire time, it never looked as it does now. Embedding methods like Word2Vec just did not exist back then. And now we have anything to vector as a simple plug and play to deep learning models.

Point clouds created from LiDAR are part of this embedding \cite{xu_2023_pointllm}. They exist as having bands detailing their values per point. Below is an example of a point cloud where the points have cartesian coordinates for 3D awareness. These coordinates are usable in any workflow both procedural and deep learning. Using LiDAR libraries we can place RGB and infrared values to each point. These types of point clouds can have up to 5 bands \cite{yang_2024_improved_tree_seg}-the coordinate, elevation, RGB colors and infrared. Then a final classification flag-such as powerlines, tree, building. 

These bands and classification can be flattened in the same way as Vision Transformers \cite{dosovitskiy_2020_an}. They are used in neural models to perform many tasks. The same is true for imagery. We take the RGB bands of the image then flatten them for a model like SAM (Segment Anything Model) \cite{dosovitskiy_2020_an}. 

\begin{figure}[h]
    \captionsetup{singlelinecheck=false, format=hang, justification=centering, labelsep=space}
    \centering
    \begin{measuredfigure}
        \includegraphics[scale=0.7]{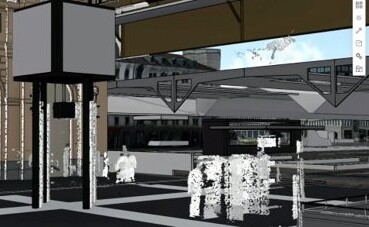}     
        \caption{Point Clouds}
    \end{measuredfigure}
\end{figure}

For the purpose of this research we convert the LiDAR values into elevation maps then subtract heights. These maps are grayscale at a low resolution with values being height of that particular point. Existing image-text to text models can interpret those values.

\section{Research Questions}
As an introduction to GeoAI we start with a procedural way of solving object height. Afterward we show modern methods. Both are reviewed and discussed. Quickly you will see what is written here appear in the coming months. Given the nature of machine learning this type of technology is inevitable.

\begin{table}[h]
\centering
\caption{Research Questions}
\begin{tabular}{|p{1.26cm} p{15.23cm}|}
\hline
\multicolumn{2}{|c|}{\textbf{Research Question}} \\ \hline
\multicolumn{1}{|l|}{\textbf{RQ1}} & How to get object heights using LiDAR and aerial imagery? \\ \hline
\multicolumn{1}{|l|}{\textbf{RQ2}} & What LiDAR and imagery models can be used in remote sensing? \\ \hline
\multicolumn{1}{|l|}{\textbf{RQ3}} & How can we get object heights using only LiDAR? \\ \hline
\end{tabular}
\end{table}

\subsection{Methods}
\begin{adjustbox}{margin=0 9 0 0}
\textbf{RQ1:} \textit{How to get object heights using LiDAR \& aerial imagery?}
\end{adjustbox}

Typically LiDAR points are classified by type. This would be trees, ground, buildings, power lines. Without this, there would have to be an intermediate step of classifying points. Point classification is a small issue and is already solved. Labelling a cluster of points as one object-that is the bigger issue. The fulcrum of the method used here is segmentation.  We combine the individual object detection by SAM and the LiDAR type to create tabular data:
\textbf{\begin{figure}[h]
    \captionsetup{singlelinecheck=false, format=hang, justification=centering, labelsep=space}
    \centering
    \begin{measuredfigure}
        \includegraphics[scale=0.2]{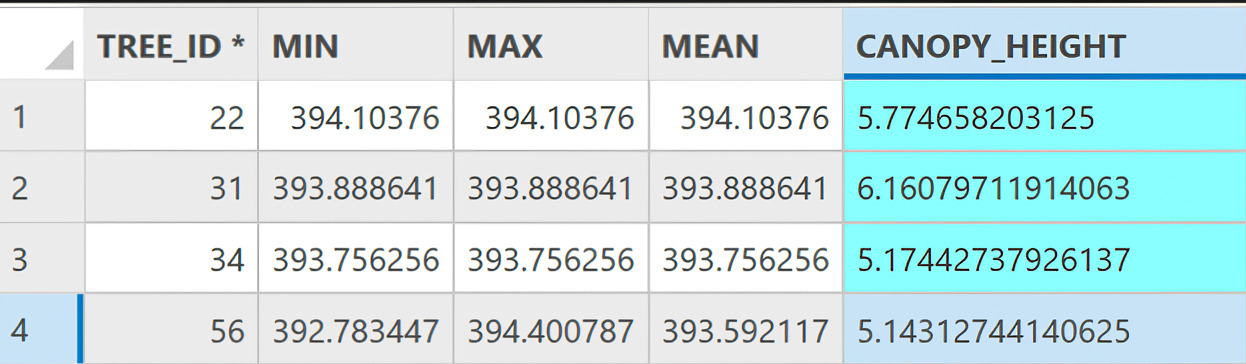}              
        \caption{Results}
    \end{measuredfigure}
\end{figure} }

The two input data, LiDAR and imagery, are processed separately then combined. Two elevation models are created, ground and object by height value. The objects are taken from the LiDAR with null outside the points. The blank white just means no data exists at those points. 

Ground elevation is blocked by the object point clouds. To remedy this we take the nearest neighboring points of missing heights to create a ground height. We then subtract object height from ground height then union the resulting differences as an object table. Once segmentation is done-the area of the objects can be placed under three metrics. Min, Max and Mean height of the object. It is feet about sea level. We include a picture of some results by object:

\begin{figure}[h]
    \captionsetup{singlelinecheck=false, format=hang, justification=centering, labelsep=space}
    \centering
    \begin{measuredfigure}
        \includegraphics[scale=0.7]{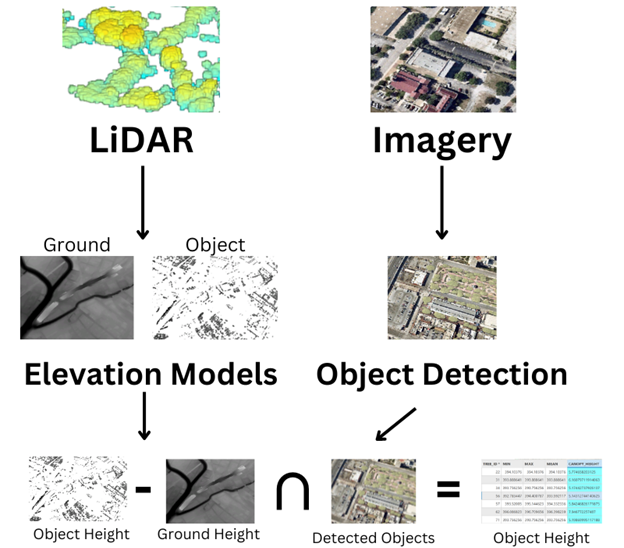}              
        \caption{Procedural Method}
    \end{measuredfigure}
\end{figure}
\begin{adjustbox}{margin=0 9 0 0}
\textbf{RQ2:} \textit{How can we get object heights using only LiDAR? }
\end{adjustbox}\\
Encoding LiDAR bands into transformer blocks is the latest trend \cite{xu_2023_pointllm}. Normal LiDAR data includes simple elevation and points classified. Given how easy it is to create custom architectures-we can repurpose past LiDAR for transformers. Most LiDAR datasets were used in convolutional neural networks and models like PointNet++ \cite{qi2017pointnetdeeplearningpoint}.

Instead of using aerial imagery and LiDAR separately-current research is finding the best results with them together. They combine the RGB colors of imagery with the LiDAR elevation, infrared-all in 5 bands. Given this combination of bands-remote sensing object heights can become the most accurate ever. 
 
\begin{figure}[h]
    \captionsetup{singlelinecheck=false, format=hang, justification=centering, labelsep=space}
    \centering
    \begin{measuredfigure}        
        \includegraphics[scale=0.8]{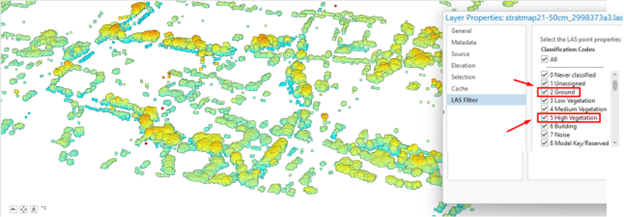}              
        \caption{LiDAR Point Classification}
    \end{measuredfigure}
\end{figure}
After adding an adapter-transformer blocks can be trained with MLP layers to do anything we want. Even multi-model LLMs which can take in LiDAR \cite{lzwlzw_2024_lzwlzwawesomeremotesensingvisionlanguagemodels}. These generative AI are similar to the original adapter for Vision Transformer. Though we flatten the new bands. Some of these models will be released to the public soon. 

\begin{adjustbox}{margin=0 9 0 0}
\textbf{RQ3:} \textit{What is the future of LiDAR and imagery processing?}
\end{adjustbox}\\
For now-using the pipeline displayed in figure RQ1 is the most accessible way to get object heights. It requires little deep learning education. Researchers can use the method in this paper with only geospatial knowledge. Machine Learning is becoming more accessible month by month to all people. The field is accelerating as it becomes easier for people to enter. 

Just a few years ago transformers only existed for NLP. Now-given how quickly vision transformers made its way into generative AI. No modal of data will be exempt from being placed into transformers and therefore LLMs. The same will be done with remote sensing generative AI. Today it is starting with ground point clouds. But soon it will turn into aerial LiDAR. We will see more remote sensing problems solved with transformers and NLP. LiDAR and imagery are no exception. Some surveys are regularly updated with the latest research on imagery and LiDAR. If you want to see the SOTA here is where you can find them:

\begin{table}[h]
\centering
\caption{Resources for Imagery \& LiDAR}
\begin{tabular}{|p{4cm}|p{12cm}|}
\hline
\multicolumn{1}{|l|}{\textbf{Focus}} & \textbf{Listing} \\ \hline
\multicolumn{1}{|l|}{\textbf{Imagery Techniques}} & \url{https://github.com/satellite-image-deep-learning/techniques} \\ \hline
\multicolumn{1}{|l|}{\textbf{LiDAR Techniques}} & \url{https://github.com/szenergy/awesome-lidar} \\ \hline
\multicolumn{1}{|l|}{\textbf{Remote Sensing LLMs}} & \begin{minipage}[t]{\linewidth}\raggedright\url{https://github.com/ZhanYang-nwpu/Awesome-Remote-Sensing-Multimodal-Large-Language-Model}\end{minipage} \\ \hline
\end{tabular}
\end{table}

\section{Limitations}
For the lifetime of object detection in vision research-accuracy of aerial imagery has been a struggle. Libraries like DeepForest \cite{weinstein_2020_deepforest}, foundational models such as SAM, do struggle to get SOTA results. For many years vision researchers have been trying to segment objects from satellite imagery. There are so many use cases for having up to date objects from satellite-it’s a hot topic. The typical result is about 70\% recall for objects today, even with SAM. For many problems and methods this is the constricting value. As such it is the constricting value for this research.

We found these libraries work in different spatial resolutions. Small 10cm resolution was good for small objects, missing all the bigger objects. 30cm was great for larger objects but missed the small objects. We took a happy medium and decided on 20cm. Notice in figure 5 the bigger tree canopies have not been detected. This is due to the 20cm resolution with SAM. It’s the perfect spatial resolution for the average tree, but not the bigger ones. In addition most of these models perform better in cities. Geography and location make a huge difference.

\begin{figure}[h]
    \captionsetup{singlelinecheck=false, format=hang, justification=centering, labelsep=space}
    \centering
    \begin{measuredfigure}
        \includegraphics[scale=0.8]{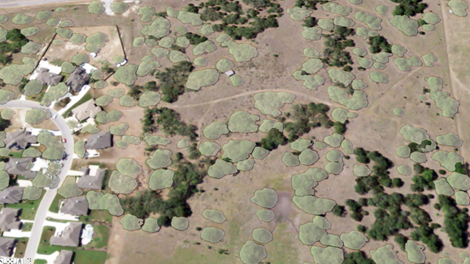}    
        \caption{SAM Results}
    \end{measuredfigure}
\end{figure}

Two solutions for this is to fine-tune SAM/DeepForest or ensemble results at different resolutions. We found lower resolutions, like 50cm, do pick up bigger object but also hallucinate objects. At 20cm there is less false positives and just miss larger objects. It is recommended by other engineers to fine-tune results to the city or rural area being inferenced. Other RGB segmentation methods exist-like in the table below. These recall numbers are average for all resolution, geography and object type.

\begin{table}[h]
\centering
\caption{RGB Detection Models}
\begin{tabular}{|p{5cm}|p{5cm}|p{5cm}|}
\hline
\multicolumn{1}{|c|}{\textbf{Models}} & \multicolumn{1}{c|}{\textbf{Recall}} & \multicolumn{1}{c|}{\textbf{Can fine-tune?}} \\ \hline
\multicolumn{1}{|l|}{\textbf{RetinaNet (Trainable)}} & 65\%+ & Yes \\ \hline
\multicolumn{1}{|l|}{\textbf{SAM (Foundational)}} & 68\% & Yes \\ \hline
\multicolumn{1}{|l|}{\textbf{DeepForest (Trees)}} & 69\% & Yes \\ \hline
\end{tabular}
\end{table}

\section{Future Research Directions}
The future of remote sensing is geospatial LLMs. The architectures are already there. Given the state of remote sensing today-there are many untried projects. The difference today is the quality of point, RGB and text embedding. The transformers architectures have become very effective. Papers like \textit{LiDAR-LLM} \cite{yang_2023_lidarllm} have promising results. The authors show increasing n-gram metrics, classification accuracies and improved sequential adherence.

\begin{figure}[h]
    \captionsetup{singlelinecheck=false, format=hang, justification=centering, labelsep=space}
    \centering
    \begin{measuredfigure}
        \includegraphics[scale=0.5]{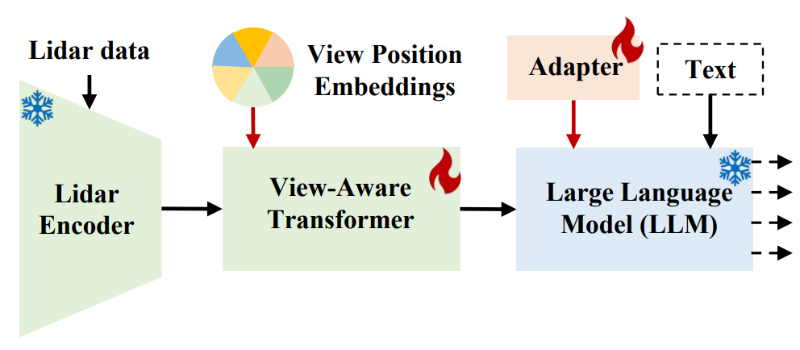}
        \caption{Example Point Cloud LLMs \cite{yang_2023_lidarllm}}
    \end{measuredfigure}
\end{figure}

In the corpus of research-more papers like these will be published soon. PointNet++, PointTransformers \cite{yang_2024_improved_tree_seg} and so many others are on the way. These transformers come in compact model scripts of pure tensor manipulation. It has become so simple to create these models due to the effectiveness of attention. For transformer models-we have not seen such effects on AI since the publication of RNNs and CNNs. 

The main issue-everything is so new. These architectures are not heavily trained. LiDAR encoders flatten the bands and classifications. The .las file is placed into an MLP and through the transformer as an embedding. With the versatility of attention and simplicity of the architecture-what we need is datasets. Datasets are the future of geospatial LLMs. We need to input geo data, text and output geo data, text. 

\section{Conclusion}
We gave an introduction to GeoAI through a procedural method for extracting object heights. The current state of remote sensing is moving from these methods to generative ones. Despite the speed of remote sensing we are not yet extracting geometries from aerial data using NLP.  The latest literature shows we are currently getting past the LiDAR embedding. But-there have not been any groundbreaking language model developments here yet. Ground LiDAR is far more popular due to its use in robotics. Research is at the cusp of creating geospatial LLMs that will turn procedural methods into NLP ones.

\newpage

\end{document}